\pdfoutput=1

\documentclass[journal]{IEEEtran}

\IEEEoverridecommandlockouts

\usepackage{mathrsfs}
\usepackage[font={small}]{caption}
\usepackage{scalerel}
\usepackage{url}
\usepackage{bbold}
\usepackage{amsfonts}
\usepackage{amsmath,amssymb,amsfonts,amsthm}
\usepackage{graphicx}
\usepackage{wrapfig}
\usepackage{mathrsfs} 
\usepackage{algorithm,algorithmic}
\usepackage{array}
\usepackage{times}
\usepackage{url}
\usepackage{subfigure}
\usepackage{cite}
\usepackage{upgreek}
\usepackage{float}
\usepackage{longtable}
\usepackage{color}
\usepackage{wasysym}
\usepackage{grffile}
\usepackage{stackengine}

\usepackage{adjustbox}
\usepackage[numbers,compress]{natbib}

\allowdisplaybreaks

\newcommand{\E}{\mathbb{E}}
\newcommand{\R}{\mathbb{R}}

\newcommand{\ab}{\mathbf{a}}

\newcommand{\gb}{\mathbf{g}}

\newcommand{\pb}{\mathbf{p}}

\newcommand{\wb}{\mathbf{w}}
\newcommand{\xb}{\mathbf{x}}

\newcommand{\zb}{\mathbf{z}}

\newcommand{\Ab}{\mathbf{A}}
\newcommand{\Bb}{\mathbf{B}}
\newcommand{\Cb}{\mathbf{C}}
\newcommand{\Db}{\mathbf{D}}

\newcommand{\Fb}{\mathbf{F}}

\newcommand{\Hb}{\mathbf{H}}
\newcommand{\Ib}{\mathbf{I}}

\newcommand{\Kb}{\mathbf{K}}

\newcommand{\Mb}{\mathbf{M}}

\newcommand{\Qb}{\mathbf{Q}}
\newcommand{\Rb}{\mathbf{R}}

\newcommand{\Zb}{\mathbf{Z}}

\newcommand{\alphab}{\boldsymbol{\alpha}}

\newcommand{\omegab}{\boldsymbol{\omega}}

\newcommand{\Ac}{\mathcal{A}}
\newcommand{\Bc}{\mathcal{B}}

\newcommand{\Hc}{\mathcal{H}}
\newcommand{\Ic}{\mathcal{I}}

\newcommand{\Vc}{\mathcal{V}}

\newcommand{\argmin}{\text{argmin}}

\newcommand{\norm}[1]{\left\lVert#1\right\rVert}
\newcommand{\tr}[1]{\text{Tr}\left[#1\right]}
\newcommand{\inn}[1]{\left<#1\right>}

\newcommand{\abs}[1]{\left|#1\right|}

\newcommand{\ex}[1]{\E\left[#1\right]}

\newtheorem{theorem}{\bf Theorem}

\newtheorem{assumption}{\bf Assumption}

\newtheorem{corollary}[theorem]{\bf Corollary}

\newtheorem{lemma}[theorem]{\bf Lemma}

\newtheorem{problem}{\bf Problem}

\newtheorem{remark}{\bf Remark}

\usepackage{etoolbox}

\title{\LARGE \bf RFN: A Random-Feature Based Newton Method for Empirical Risk Minimization in Reproducing Kernel Hilbert Spaces}

\author{Ting-Jui Chang and Shahin Shahrampour, {\it Senior Member}, {\it IEEE}  
\thanks{T.J. Chang and S. Shahrampour are with the Department of Mechanical and Industrial Engineering, Northeastern University, Boston, MA 02115, USA. 
{\tt\footnotesize email:\{chang.tin,s.shahrampour\}@northeastern.edu}.}
}

\begin{document}

\maketitle
\thispagestyle{plain}
\pagestyle{plain}

%%%%%%%%%%%%%%%%%%%%%%%%%%%%%%%%%%%%%%%%%%%%%%%%%%%%%%%%%%%%%%%%%%%%%%%%%%%%%%%%
\begin{abstract}
In supervised learning using kernel methods, we often encounter a large-scale finite-sum minimization over a reproducing kernel Hilbert space (RKHS). Large-scale finite-sum problems can be solved using efficient variants of Newton method, where the Hessian is approximated via sub-samples of data. In RKHS, however, the dependence of the penalty function to kernel makes standard sub-sampling approaches inapplicable, since the gram matrix is not readily available in a low-rank form. In this paper, we observe that for this class of problems, one can naturally use kernel approximation to speed up the Newton method. Focusing on randomized features for kernel approximation, we provide a novel second-order algorithm that enjoys local superlinear convergence and global linear convergence (with high probability). We derive the theoretical lower bound for the number of random features required for the approximated Hessian to be close to the true Hessian in the norm sense. Our numerical experiments on real-world data verify the efficiency of our method compared to several benchmarks. 
\end{abstract}

\section{Introduction}
At the heart of many supervised machine learning problems, a learner must solve the following risk minimization 
\begin{align}\label{eq:risk}
    \min_{\wb\in \R^d} \left\{ F(\wb)\triangleq \frac{1}{n}\sum\limits_{i=1}^n\ell\big(y_i,f(\xb_i;\wb)\big)+\lambda R(\wb)\right\},
\end{align}
where $\{(\xb_i,y_i)\}_{i=1}^n$ are input-output data samples generated independently from an unknown distribution, $\ell$ is a task-dependent loss function, and $R$ is a regularizer. Furthermore, $f$ is a certain function class, parameterized by $\wb$, on which the learner wants to minimize the risk. As an example, for linear models we simply have $f(\xb;\wb)=\xb^\top\wb$. 

First-order optimization algorithms have been widely used to solve large-scale optimization problems of form \eqref{eq:risk} (see \cite{bottou2018optimization} for a recent survey). Relying solely on the gradient information, these methods converge to (local) optima. However, second-order algorithms employ the curvature information to properly re-scale the gradient, resulting in more appropriate directions and much faster convergence rates. As an example, in the unconstrained optimization, Newton method pre-multiplies the gradient by the Hessian inverse at each iteration. It is quite well-known that under some technical assumptions, Newton’s method can achieve a locally super-linear convergence rate for strongly convex problems (see e.g., Theorem 1.2.5 in \cite{nesterov1998introductory}). However, the cost of Hessian inversion is the major drawback of Newton’s method in practice.

To improve the (per iteration) time complexity, various approaches have been explored in the literature for approximately capturing the Hessian information. Popular methods in this direction include sub-sampling the Hessian matrix \cite{bollapragada2018exact, roosta2019sub, xu2016sub}, sketching techniques \cite{pilanci2017newton}, as well as quasi-Newton methods \cite{broyden1970convergence, fletcher1970new, goldfarb1970family, shanno1970conditioning} and its stochastic variants \cite{schraudolph2007stochastic, mokhtari2014res, byrd2016stochastic, moritz2016linearly, bordes2009sgd,sohl2014fast, mokhtari2015global, mokhtari2018iqn, gower2016stochastic, zhao2018stochastic, chang2019accelerated, zhang2021faster}. 

Nevertheless, when the function class of $f$ in \eqref{eq:risk} is a reproducing kernel Hilbert space (RKHS), due to the special structure of the problem, some of the aforementioned methods are not directly applicable and need some adjustments. 

\subsection{Risk Minimization in RKHS}
In this paper, we restrict our attention to risk minimization in the case that the function $f$ in \eqref{eq:risk} belongs to a RKHS. In particular, consider a symmetric function $k(\cdot,\cdot)$ such that 
$$\sum^n_{i,j=1}\alpha_i \alpha_j k(\xb_i,\xb_j)\geq 0,$$ for $\alphab=[\alpha_1,\ldots,\alpha_n]^\top\in \R^n$. Then, $k(\cdot,\cdot)$ is called a positive (semi-)definite kernel and can define a Hilbert space $\Hc$ where $f(\xb;\wb)=\sum_{i=1}^nw_ik(\xb,\xb_i)$. This class of functions forms the basis of kernel methods that are powerful tools for data representation and are commonly used in machine learning and signal processing \cite{shawe2004kernel,perez2004kernel}. In this scenario, the objective function in \eqref{eq:risk} takes the following form
\begin{align}\label{eq:riskrkhs}
     F(\wb)= \frac{1}{n}\sum\limits_{i=1}^n\ell\Big(y_i,\sum_{j=1}^nw_jk(\xb_i,\xb_j)\Big)+\frac{\lambda}{2} \norm{f}^2_{\Hc},
\end{align}
where the regularizer in \eqref{eq:risk} is the RKHS norm. Let us denote the kernel (gram) matrix as $[\Kb]_{ij}=k(\xb_i,\xb_j)$. The definition of inner product in RKHS immediately implies that $\norm{f}^2_{\Hc}=\wb^\top\Kb\wb$ (see e.g., page 62 of \cite{shawe2004kernel}). Then, assuming that $\ell$ is twice-differentiable, the Hessian of the objective function in \eqref{eq:riskrkhs} can be calculated as follows 
\begin{equation}\label{eq:hessian}
    \Hb (\wb) \triangleq \nabla^2 F(\wb)= \frac{1}{n}\Kb\Db(\wb)\Kb + \lambda \Kb,
\end{equation}
where $\Db(\wb)\in \R^{n\times n}$ is a diagonal matrix defined as
\begin{align}
    [\Db(\wb)]_{ii}=\ell^{\prime\prime}\Big(y_i,\sum_{j=1}^nw_jk(\xb_i,\xb_j)\Big).
\end{align}
The inversion of the Hessian matrix requires an order of $n^3$ operations, which is costly. On the other hand, observe that in \eqref{eq:hessian}, the diagonal structure of $\Db(\wb)$ and the symmetry of $\Kb$ together imply that $\Kb\Db(\wb)\Kb$ can be trivially represented as a sum of rank-one matrices. However, $\Kb$, which appears as a result of the regularization term, may be dense and not readily available in a low-rank form. In other words, decomposing $\Kb$ to a low-rank matrix also requires effort, so we cannot directly apply sub-sampling Newton techniques, such as those in \cite{xu2016sub,bollapragada2018exact,roosta2019sub} to optimize the objective function \eqref{eq:riskrkhs}. This naturally raises the following question, which we pursue in this paper:
\begin{problem}\label{P1}
Given the explicit connection of the Hessian \eqref{eq:hessian} to the gram matrix $\Kb$, can we use kernel approximation techniques to improve the per iteration time complexity of the Newton method?
\end{problem}

\subsection{Our Contributions}
In this paper, we answer to Problem \ref{P1} in the affirmative by providing the following contributions:
\begin{itemize}
    \item We apply the idea of randomized features for kernel approximation \cite{rahimi2008random} to approximate the Hessian \eqref{eq:hessian}. Our algorithm is thus dubbed Random-Feature based Newton (RFN). The detailed derivation of RFN is explained in Section \ref{sec:alg}.
    \item The key to our technical analysis is Lemma \ref{spectral_lemma}, which shows that when enough random features are sampled, the approximate Hessian is close to $\Hb(\wb)$ in the spectral norm sense. Our analysis relies on matrix concentration inequalities and explicitly derives the theoretical lower bound of the number of random features to get $\epsilon$-close to the Hessian.
    \item We prove that RFN enjoys local superlinear convergence and global linear convergence in the high probability sense (Section \ref{sec:theory}). 
    \item Our numerical experiments on real-world datasets (Section \ref{sec:simul}) provide a performance comparison among RFN, classical Newton, L-BFGS, and a variation of sub-sampled Newton methods (as they are not directly applicable to \eqref{eq:riskrkhs}). We illustrate that RFN achieves a superior loss vs. run-time rate against its competitors. %We also investigate the sensitivity of RFN with respect to a number of hyper-parameters.  
\end{itemize}
We have included the omitted proofs in the Appendix (Section \ref{sec:appendix}).

\subsection{Related Literature}
Inspired by the success of stochastic first-order algorithms for large-scale data analytics, the stochastic forms of second-order optimization have received more attention in the recent literature. In this section, we review several stochastic second-order methods, and we split them into two categories: quasi-Newton methods and second-order Hessian-based methods.

\noindent
\textbf{-- Quasi-Newton Methods:}
Instead of performing the expensive computation of the Newton step (which involves Hessian inversion), quasi-Newton methods approximate the Hessian by using the information obtained from gradient evaluations. BFGS algorithm \cite{broyden1970convergence, fletcher1970new, goldfarb1970family, shanno1970conditioning} is a seminal work of this type. Recent works \cite{schraudolph2007stochastic, mokhtari2014res, byrd2016stochastic, moritz2016linearly, bordes2009sgd,sohl2014fast, mokhtari2015global, mokhtari2018iqn, gower2016stochastic, zhao2018stochastic, chang2019accelerated, zhang2021faster} in this area have focused on stochastic quasi-Newton methods to obtain curvature information in a computationally inexpensive manner. \citet{mokhtari2014res} propose a stochastic regularized version of BFGS algorithm to avoid the problem of singular curvature estimates. \citet{byrd2016stochastic} %, instead of using difference of stochastic gradients, 
apply sub-sampled Hessian-vector products to stabilize the curvature estimation based on a stochastic limited memory BFGS (L-BFGS) algorithm.  Another advancement in this line of work is provided by \citet{moritz2016linearly} who propose an algorithm based on L-BFGS by incorporating ideas from stochastic variance-reduced gradient (SVRG) to achieve linear convergence to the optimum. \citet{gower2016stochastic} employ the idea of SVRG to propose a stochastic block BFGS algorithm using sketching techniques and show practical speed-ups for common machine learning problems. Based on the algorithm structure of \cite{moritz2016linearly}, \citet{zhao2018stochastic} propose a coordinate transformation framework to analyze stochastic L-BFGS type algorithms and present improved convergence rates and computational complexities.  Using momentum for L-BFGS \citet{chang2019accelerated} prove an accelerated linear convergence rate with better dependence
on the condition number. For non-convex optimization, \citet{zhang2021faster} propose a novel stochastic quasi-Newton method with an improved stochastic first-order oracle complexity for reaching an $\epsilon$ first-order stationary point.

\noindent
\textbf{-- Second-Order Hessian based Methods:}
The appealing feature of Newton method is its fast local convergence rate. However, there are two main issues for the implementation of the classical Newton method: the cost of Hessian construction, and the cost of Hessian inversion. For example, in application to the family of generalized linear models (GLMs) involving an $n\times d$ data matrix, the computation of the full Hessian costs $O(nd^2)$ and the matrix inversion takes $O(d^3)$ time. This high cost, especially for large-scale applications, has motivated researchers to apply randomization techniques, and thus sub-sampled Newton methods have gained a good deal of attention recently. 

In \cite{byrd2011use, byrd2012sample}, the authors establish the convergence of the modified Newton method with sub-sampled Hessian. %\cite{byrd2012sample} also provides work-complexity bounds of their method and reports the experimental results with a sub-sampled Newton-CG method. 
Under a similar setting to \cite{byrd2011use, byrd2012sample}, \citet{wang2015subsampled} provide  modifications in order to get better estimated Hessian and time cost performance. Within the context of deep neural networks, \citet{martens2010deep} proposes a sub-sampled Gauss-Newton method for the training and studies the choice of the regularization parameter. 

\citet{Erdogdu2015ConvergenceRO} propose a Newton-like method, where the Hessian is approximated by sub-sampling the true Hessian and computing its truncated eigenvalue decomposition. Their work establishes non-asymptotic local convergence rates for the uniform sub-sampling of the Hessian. \citet{pilanci2017newton} propose another Hessian approximation method, called Newton sketching, which approximates the true Hessian through random projection matrices. This method is applicable for the case where the square root of the full Hessian is available, and the best complexity results are achieved when the randomized Hadamard transform is used.

Authors of \cite{bollapragada2018exact, roosta2019sub} analyze the global and local convergence rates for sub-sampled Newton methods with different sampling rates for gradient and Hessian approximations. \citet{bollapragada2018exact} show the convergence results in expectation, whereas \cite{roosta2019sub} provides high probability guarantees by applying matrix concentration inequalities \cite{tropp2010computational, tropp2015introduction}. The work of \cite{roosta2019sub} further relaxes a common assumption in the literature: though the objective function is assumed to be strongly convex, the individual functions are only weakly convex. Along this line of works, \citet{xu2016sub} build the approximated Hessian by applying non-uniform sampling based on the data matrix to get better dependence on problem specific quantities.

\citet{agarwal2017second} propose a method to compute an unbiased estimator of the inverse Hessian based on the power expansion of the Hessian inverse. The method achieves a time complexity scaling linearly with the size of variables. This is followed by an improved and simplified convergence analysis in \cite{mutny2016stochastic}.

The main distinction of our work with the literature is that we consider risk minimization in RKHS, where there are explicit connections between the Hessian and the gram matrix. We leverage this fact to approximate the Hessian, and we provide theoretical guarantees for the RFN method. Our prior work \cite{chang2020global} studied only the global convergence of RFN for simple kernels, but this work provides a comprehensive analysis on both global and local convergence rates for {\it composite} kernels explained in Section \ref{sec:composite}. This allows us to achieve sharper theoretical lower bound for the number of random features required for convergence. Furthermore, we illustrate the efficiency of RFN compared to other benchmarks on real-world datasets.

\section{Random-Feature Based Newton Method}\label{sec:alg}
\subsection{Notation}
We denote by $\tr{\cdot}$ the trace operator, by $\inn{\cdot,\cdot}$ the standard inner product, by $\norm{\cdot}$ the spectral (respectively, Euclidean) norm of a matrix (respectively, vector), by $O(\cdot)$ (respectively, $\Omega(\cdot)$) the Big O (respectively, Big Omega) notation in complexity theory, and by $\ex{\cdot}$ the expectation operator. Boldface lowercase variables (e.g., $\ab$) are used for vectors, and boldface uppercase variables (e.g., $\Ab$) are used for matrices. $[\Ab]_{ij}$ denotes the $ij$-th entry of matrix $\Ab$. $\lambda_{\min}(\Ab)$ (respectively, $\lambda_{\max}(\Ab)$) denotes the smallest (respectively, largest) eigenvalue of matrix $\Ab$. The symbol ``$\preceq$" is used for matrix inequality and $\Ab \preceq \Bb$ implies that the matrix $(\Bb-\Ab)$ is positive semi-definite. $|\Ic|$ represents the cardinality of the set $\Ic$. The vectors are all in column form. 

\subsection{Background on Random Features for Kernel Approximation}
As discussed in the introduction, the Hessian of the objective function in \eqref{eq:riskrkhs} can be written as $\Hb(\wb)=\frac{1}{n}\Kb\Db(\wb)\Kb + \lambda \Kb$, where $[\Kb]_{ij}=k(\xb_i,\xb_j)$ is the gram matrix. The Hessian is a square matrix of size $n$, and a plain inversion of that in Newton method introduces a prohibitive cost of $O(n^3)$. As $n$ is the number of data points, this cost is specifically expensive for big data problems. 

To find a low-rank representation of Hesssian, the key is to approximate the gram matrix $\Kb\in \R^{n\times n}$, which in general is dense. An elegant method for kernel approximation, called random Fourier features, is introduced by \citet{rahimi2008random}. Let $p(\omegab)$ be a probability density with support $\Omega \subseteq \R^d$. Consider any kernel function with the following integral form
\begin{equation}\label{kernel function}
    k(\xb,\xb') = \int_\Omega \phi(\xb,\omegab)\phi(\xb',\omegab)p(\omegab)d\omegab,
\end{equation}
where $\phi(\xb,\omegab) : \R^d \rightarrow \R$ is a feature map. We can immediately see from \eqref{kernel function} that the kernel function can be approximated via Monte-Carlo sampling as
\begin{equation}\label{kernel aprroximation}
    k(\xb,\xb') \approx \frac{1}{m} \sum_{s=1}^m \phi(\xb,\omegab_s)\phi(\xb',\omegab_s),
\end{equation}
where $\{\omegab_s\}_{s=1}^m$ are independent samples from the density $p(\omegab)$ and are called {\it random features}. There exist many kernels taking the form \eqref{kernel function}, including shift-invariant kernels \cite{rahimi2008random} and dot product (e.g., polynomial) kernels \cite{kar2012random} (see Table 1 in \cite{yang2014random} for an exhaustive list). Gaussian kernel, for example, can be approximated using $\phi(\xb,\omegab)=\sqrt{2}\cos(\omegab^\top \xb+b)$ where $\omegab$ follows a Gaussian distribution and $b$ has a uniform distribution on $[0,2\pi]$. It is common to assume that the feature map is uniformly bounded (as evident in the case of cosine). For simplicity, we assume $\abs{\phi(\xb,\omegab)}\leq 1$ for any $\xb,\wb\in \R^d$. Let us now define
%We now define for the distribution-feature map pair $\big(p(\cdot),\phi(\cdot,\cdot)\big)$
\begin{align}\label{zvector}
\zb(\omegab)\triangleq[\phi(\xb_1,\omegab),\ldots,\phi(\xb_n,\omegab)]^\top.
\end{align}
Then, based on \eqref{kernel aprroximation}, the gram matrix $\Kb$ can be approximated with $\Zb\Zb^\top$ where $\Zb\in \R^{n\times m}$ is the following matrix
\begin{align}\label{trasformed matrix}
\Zb &\triangleq \frac{1}{\sqrt{m}}[\zb(\omegab_1),\ldots,\zb(\omegab_m)].
\end{align}
When $m<n$ in above, $\Zb\Zb^\top$ has a lower rank than $n$, saving computational cost when used to find the Newton direction. However, unlike the finite sum problem considered in \cite{roosta2019sub}, where $n$ (\# of data points) can be much greater than $d$ (the size of parameters to learn), in the optimization problem for RKHS \eqref{eq:riskrkhs} the decision variable is also of the same size as $n$. Therefore, only replacing $\Kb$ by a low-rank form $\Zb\Zb^{\top}$ leads to an approximated Hessian that is singular. Thus, we only focus on positive definite kernels of composite type, as described next.  

\subsection{Composite Kernel Function}\label{sec:composite}
We consider gram matrices built on a composite kernel with the following form:
\begin{equation}\label{eq: composite kernel function}
    k(\xb,\xb^{\prime}) = k_1(\xb,\xb^{\prime}) + k_2(\xb,\xb^{\prime}),
\end{equation}
where $k_1(\xb,\xb^{\prime})$ can be expressed as \eqref{kernel function}, and $k_2(\xb,\xb^{\prime})=\mu \mathbb{1}(\xb=\xb')$, where $\mathbb{1}$ is the indicator function. From this definition, we have that $\Kb = \Kb_1 + \mu\Ib$, and the random feature method is applied to approximate $\Kb_1$. Note that this expression is not restrictive, because for any positive-definite kernel $\Kb$, we can always find a positive semi-definite kernel $\Kb_1$ and a parameter $\mu$, such that \eqref{eq: composite kernel function} holds. The expression only allows us to streamline our presentation.

\subsection{Algorithm: Random-Feature Based Newton (RFN)}
RFN leverages random feature method for Hessian approximation to execute the Newton method. Following \eqref{eq:hessian} and using the approximation $\Kb\approx\Zb\Zb^\top + \mu\Ib$, where $\Zb$ is given in \eqref{trasformed matrix}, we derive the following approximated Hessian
\begin{equation}\label{eq:hessianAprox}
    \widehat{\Hb}(\wb)= \frac{1}{n}(\Zb\Zb^\top+\mu\Ib)\Db(\wb)(\Zb\Zb^\top+\mu\Ib) + \lambda (\Zb\Zb^\top+\mu\Ib).
\end{equation}
Define $\Cb\triangleq (\Zb\Zb^{\top}+\mu\Ib)$. By matrix inversion lemma, $\widehat{\Hb}^{-1}(\wb)$ can be written as
\begin{equation}\label{eq:hessianIL1}
    \widehat{\Hb}^{-1}(\wb) = \frac{\Cb^{-1} - \big(\lambda n\Db^{-1}(\wb) + \Cb\big)^{-1}}{\lambda},
\end{equation}
where we can observe that both $\Cb$ and $\big(\lambda n\Db^{-1}(\wb) + \Cb\big)$ are expressed as the sum of a low-rank matrix and a diagonal matrix. %, which can also be converted in a computationally efficient form by matrix inversion lemma again. 
Therefore, denoting the diagonal matrix $\Db_{\mu}(\wb)\triangleq(\lambda n\Db^{-1}(\wb) + \mu\Ib)$ and substituting $\Cb= (\Zb\Zb^{\top}+\mu\Ib)$ in above, we have that
\begin{equation}\label{eq:hessianIL2}
\begin{split}
    &\widehat{\Hb}^{-1}(\wb)\\
    = &\frac{1}{\lambda}\bigg[ \bigg. \frac{1}{\mu}\left[\Ib - \Zb(\mu\Ib + \Zb^{\top}\Zb)^{-1}\Zb^{\top}\right]\\
    + &\Db_{\mu}^{-1}(\wb) - \Db_{\mu}^{-1}(\wb)\Zb\big(\Ib + \Zb^{\top}\Db_{\mu}^{-1}(\wb)\Zb\big)^{-1}\Zb^{\top}\Db_{\mu}^{-1}(\wb) \bigg.\bigg],
\end{split}    
\end{equation}
which can reduce the time complexity of computing $\widehat{\Hb}^{-1}(\wb)\nabla F(\wb)$ from $O(n^3)$ to $O(m^2n+m^3)$. Then, we can use $\widehat{\Hb}^{-1}(\wb)$ to perform a computationally efficient Newton update, as described in Algorithm \ref{alg:RFN}.  
% It is immediate that when $m<n$, the above approximated hessian is singular, so in order to make $\widehat{\Hb}(\wb)$ invertible, a regularization term $\mu \Ib$ (for some $\mu>0$) is added to the above, and the new approximate can be written as,
% \begin{equation}\label{eq:hessianAproxmu}
%     \widehat{\Hb}_\mu(\wb)\triangleq\frac{1}{n}\Zb\left[\Zb^\top\Db(\wb)\Zb + \lambda \Ib\right]\Zb^\top + \mu \Ib.
% \end{equation}
% Let $\Cb(\wb)\triangleq\Zb^\top\Db(\wb)\Zb + \lambda \Ib$. By matrix inversion lemma, $\widehat{\Hb}^{-1}_\mu(\wb)$ can be written as 
% \begin{align}\label{eq:hessianIL}
%     \widehat{\Hb}^{-1}_\mu(\wb)=\frac{1}{\mu}\left[\Ib - \Zb\Big(n\mu \Cb^{-1}(\wb)+\Zb^\top\Zb\Big)^{-1}\Zb^\top\right],
% \end{align}
% which can reduce the time complexity of computing $\widehat{\Hb}^{-1}_\mu(\wb)\nabla F(\wb)$ from $O(n^3)$ to $O(m^2n+m^3)$. Then, we can use $\widehat{\Hb}^{-1}_\mu(\wb)$ to perform a computationally efficient Newton update, as described in Algorithm \ref{alg:RFN}.  

\begin{algorithm}[tb]
   \caption{Random-Feature Based Newton (RFN)}
   \label{alg:RFN}
\begin{algorithmic}[1]
   \STATE {\bfseries Input:} Initial point $\wb_0$, \# of iterations $t_0$, \# of random features $m$, density function of random features $p(\omegab)$, feature map $\phi$, hyper-parameters for backtracking line search, regularization parameters $\lambda,\mu>0$.
   \STATE {\bfseries Output:} $\wb_{t_0}$
  
   \FOR{$t=0$ {\bfseries to} $t_0-1$}
   \STATE Sample $m$ independent random features from $p(\omegab)$ and construct the feature matrix $\Zb$ as in \eqref{trasformed matrix}.
   \STATE Compute $\gb(\wb_t)=\nabla F(\wb_t)$, the gradient of the objective \eqref{eq:riskrkhs}.
   \STATE Compute $\widehat{\Hb}^{-1}(\wb_t)$ as given in \eqref{eq:hessianIL2}.
   \STATE Update $\wb_{t+1}=\wb_t - \alpha_t \widehat{\Hb}^{-1}(\wb_t)\gb(\wb_t)$, where $\alpha_t$ is selected via backtracking line search.
   \ENDFOR

\end{algorithmic}
\end{algorithm}

\subsection{Adjustment of Sub-sampled Newton Methods}\label{sec:ssn}
We now revisit sub-sampled Newton methods to elaborate on the differences between RFN and these methods. Sub-sampled Newton algorithms often end up working with Hessians of the form  
$$\Hb(\wb)=\frac{1}{n}\sum_{i=1}^n\ab_i(\wb)\ab^\top_i(\wb)+\Qb(\wb),$$ where $\ab_i(\wb)$ is a vector and $\Qb(\wb)$ is a data-independent matrix (see e.g., \cite{xu2016sub}). Given this structure, if we randomly select a subset $\Ic$ of data points, the Hessian can be easily approximated via 
$$\widehat{\Hb}(\wb) = \frac{1}{|\Ic|}\sum_{i\in \Ic}\ab_i(\wb)\ab^\top_i(\wb)+\Qb(\wb).$$
This would reduce the time cost of Hessian construction by a factor of $|\Ic|/n$. Furthermore, since the approximated Hessian still consists of sum of rank-one matrices (in the data-related part), one can apply conjugate gradient (CG) method when computing the newton step to further speed up the process \cite{xu2016sub,roosta2019sub,bollapragada2018exact}.

%One common aspect of many empirical risk minimization (ERM) problems in machine learning is that in equation \eqref{eq:risk}, the loss function  $\ell(y_i, f(\xb_i;\wb))$ is of the form: $\ell(y_i, \xb_i^\top\wb)$, which is the so-called generalized linear model (GLM), and an $\ell_2$ norm is often chosen as the regularizer $R(\wb)$. Given a loss function form like this, 
%the hessian of this problem can be written as  \cite{xu2016sub}. 

 However, it turns out that the Hessian of the  objective function \eqref{eq:riskrkhs}, i.e., $\frac{1}{n}\Kb\Db(\wb)\Kb + \lambda \Kb$, cannot be directly handled by sub-sampling, since the regularizer in this case is indeed {\it data-dependent}, and writing it as a sum of rank-one matrices requires a low-rank decomposition of $\Kb$. In this case, sub-sampled Newton techniques can be adjusted using the Nystr{\"o}m method for column sampling (see e.g., \cite{gittens2016revisiting} for a review on Nystr{\"o}m method). Let $\Vc=\{1,\ldots,n\}$ and $\Ic$ denote a (random) subset of $\Vc$. Furthermore, denote by $\Kb(\Ac,\Bc)$ the sub-matrix of $\Kb$ with rows in $\Ac\subseteq \Vc$ and columns in $\Bc\subseteq\Vc$. Then, the gram matrix $\Kb$ can be approximated as 
 \begin{align}\label{eq:approx1}
     \Kb\approx \Kb_1(\Vc,\Ic)\Kb_1(\Ic,\Ic)^{\dagger}\Kb_1(\Ic,\Vc) + \mu\Ib,
 \end{align}
 where $^{\dagger}$ denotes the pseudo-inverse. %This decomposition introduces  $O(n|\Ic|^2)$ time cost (if we apply CG method) but circumvents the need for matrix inversion. 
 Notice that the first term of the Hessian can be trivially sub-sampled since
 \begin{align}\label{eq:approx2}
     \Kb\Db(\wb)\Kb=\sum_{i=1}^n[\Db(\wb)]_{ii}\Kb(\Vc,i)\Kb(i,\Vc).
 \end{align}
 In the similar spirit as  \cite{xu2016sub,bollapragada2018exact,roosta2019sub}, we call this algorithm SSNCG, as it is a sub-sampled Newton method, where CG is used to find the Newton step.
 
\noindent 
\begin{remark}
Since the focus of this work is on random features, we only compare RFN to the case that $\Ic$ is chosen uniformly at random, i.e., the method would be a variant of uniform sub-sampled Newton \cite{roosta2019sub}. Nonuniform sampling methods give better approximations of the kernel at the cost of modifying the uniform sampling distribution. We refer the reader to Table 1 in \cite{gittens2016revisiting} for various guarantees on the approximation quality via different sampling schemes.  
\end{remark}

\subsection{Comparison of Time Complexity}
Before stating our main results, we present the time complexity of finding the Newton step for four methods: Newton method, sub-sampled Newton method solved with CG exactly, sub-sampled Newton method solved with CG inexactly up to $\varepsilon$ error, and RFN. The number of random features used by RFN is denoted by $m$. The size of the sub-sampled data points is $|\Ic|$, as discussed in Section \ref{sec:ssn}. $\kappa_{SSN}$ represents the upper bound of the condition number of the Hessian generated from sub-sampled data. Note that for each method, the reported complexity excludes the cost of obtaining the gradient, because that cost is the same for all methods.
\begin{table}[h!]
\caption{Time complexity per iteration. ($n$: \# of data points. $\Ic$: the subsampled index set. $m$: \# of sampled random features.)}
\centering
\begin{adjustbox}{width=1\columnwidth}
\begin{tabular}{||c c||} 
 \hline
 Method & Complexity per iteration\\ [0.5ex] 
 \hline\hline
 Newton & $O(n^3)$ \\ 
 SSNCG (exact) & $O(|\Ic|n^2)$ \\
 SSNCG (inexact) & $O(|\Ic|^2n + |\Ic|n\sqrt{\kappa_{SSN}}\log\varepsilon^{-1})$ \\
 RFN & $O(m^2n+m^3)$ \\
 \hline
\end{tabular}
\end{adjustbox}
\label{table:complexity}
\end{table}

\section{Theoretical Results}\label{sec:theory}
In this section, we study the convergence properties of RFN. In order to establish our results, we need to prove that the approximated Hessian mimics the original Hessian with high probability, which is shown in Lemma \ref{spectral_lemma}. Then, we can show the global and local convergence of RFN in Sections \ref{sec:global} and \ref{sec:local}, respectively.

%Notice that, throughout, for the presentation of our theoretical results we work with $\widehat{\Hb}(\wb)$ in  \eqref{eq:hessianAprox} (rather than $\widehat{\Hb}_\mu(\wb)$). The main reason is that $\widehat{\Hb}(\wb)$ is the approximate to the original hessian, whereas $\widehat{\Hb}_\mu(\wb)$ is used for implementation purposes. We will later see the practical effect of $\mu$ in the numerical experiments (Section \ref{sec:simul}).

\subsection{Norm Bound for Hessian Approximation}
Throughout the paper, we adhere to the following assumptions:
\begin{assumption}\label{A1} (Bounded Eigenvalues of Hessian)
The objective function $F$ is twice-differentiable, $\gamma$-strongly convex, and $L$-smooth. The smallest and largest eigenvalues of the Hessian are bounded as follows
    \begin{equation}\label{eq:strong_convexity}
        \gamma\Ib \preceq \Hb(\wb)\preceq L\Ib,~~~\forall{\wb}\in \R^n,
    \end{equation}
    and also $\Db(\wb)$ satisfies
    \begin{equation}\label{upperbound_D}
        \|\Db(\wb)\|\leq\lambda_1<\infty, ~~~\forall{\wb}\in \R^n.
    \end{equation}
    %For an approximated hessian using $m$ random features, we denote its maximum and minimum eigenvalues as $\lambda_{\max}(\widehat{\Hb}(\wb))$ and $\lambda_{\min}(\widehat{\Hb}(\wb))$ respectively.
\end{assumption}
Condition \eqref{upperbound_D} is satisfied for common loss functions (e.g., quadratic loss and logistic loss). Furthermore, if we solve the risk minimization \eqref{eq:riskrkhs} with a positive definite kernel, the gram matrix $\Kb$ would be positive definite, and therefore, Assumption \ref{A1} is satisfied. We also define the condition number $\kappa$ of the Hessian matrix as follows: 
\begin{equation} \label{eq:kappa}
        \kappa\triangleq \frac{L}{\gamma}. \\
    \end{equation}

\noindent
\begin{assumption}\label{A2}(Lipschitz Continuity of Hessian)
The Hessian of the objective function $F$ is Lipschitz continuous, i.e., there exists a constant $M > 0$, such that
    \begin{equation}\label{eq:lips}
        \norm{\Hb(\wb_1)-\Hb(\wb_2)}\leq M\norm{\wb_1-\wb_2}, \forall{\wb_1, \wb_2} \in \R^n.
    \end{equation}
\end{assumption}
Assumption \ref{A2} is commonly used in the literature to establish local superlinear convergence of sub-sampled Newton methods (see e.g., \cite{bollapragada2018exact, roosta2019sub}). The assumption is used to prove the local superlinear convergence of the original Newton method as well (see e.g., Theorem 1.2.5 in \cite{nesterov1998introductory}).

We are now ready to show that with high probability the approximated Hessian is close enough to the original Hessian when large enough random features are sampled.

\begin{lemma}\label{spectral_lemma} (Spectrum Preserving Inequality) Suppose Assumption \ref{A1} holds. Define $$\zeta \triangleq \lambda + \lambda_1\norm{\frac{\Zb\Zb^{\top}+\mu\Ib}{n}} + \lambda_1\norm{\frac{\Kb_1+\mu\Ib}{n}}.$$ For $0<\delta<1$,
% Suppose \eqref{upperbound_D} holds.
% Let $\Tilde{\tau}:\R^d \rightarrow \R$ be a measurable function such that $\Tilde{\tau}(\omegab)\geq \tau_{\lambda^{\prime}}(\omegab)$ for all $\omegab\in \R^d$, and assume 
% $s_{\Tilde{\tau}}$ (defined in \eqref{eq: integral of the rescaled pdf}) to be finite.
% Let $\delta \in (0,1)$ be fixed constants and $\epsilon\in(0,\zeta/2]$, where $\zeta\triangleq 2\left(\lambda + \lambda_1\norm{\frac{\Zb\Zb^{\top}}{n}} + \lambda_1\norm{\frac{\Kb}{n}}\right)\norm{\Kb}$. Assume that $||\Kb||_2 \geq \lambda^{\prime}$.
if $$m = \Omega\left(\frac{\zeta^2n\norm{\Kb_1}}{\epsilon^2\gamma^2}\log\frac{\text{Tr}(\Kb_1)}{\norm{\Kb_1}\delta}\right),$$
random features are sampled from $p(\omegab)$, we have that 
$$\text{Pr}\left(\norm{\widehat{\Hb}(\wb)-\Hb (\wb)}\leq \epsilon\gamma\right)\geq 1-\delta.$$
\end{lemma}

\begin{corollary}\label{Corollary: upperlower approximated hessian}
Under assumptions of Lemma \ref{spectral_lemma}, with probability at least $1-\delta$, we have
\begin{align}\label{eq:upperlower}
    [(1-\epsilon)\gamma]\cdot\Ib\preceq\widehat{\Hb}(\wb)\preceq[(1+\epsilon)L]\cdot\Ib,
\end{align}
where $\widehat{\Hb}(\wb)$ is ensured to be positive-definite when $\epsilon < 1$.

\begin{proof}
From Lemma \ref{spectral_lemma}, we have $\norm{\widehat{\Hb}(\wb)-\Hb (\wb)}\leq \epsilon\gamma$, which implies the following relationship
\scalebox{.95}{\parbox{\columnwidth}{
\begin{equation}\label{eq: upperlowerdeduction}
\begin{split}
    &\Hb(\wb) - \epsilon\gamma\cdot\Ib \preceq\widehat{\Hb}(\wb)\preceq\Hb(\wb) + \epsilon\gamma\cdot\Ib\\
    \implies& (1-\epsilon)\Hb(\wb)\preceq \widehat{\Hb}(\wb)\preceq (1+\epsilon)\Hb(\wb)\\
    \implies& [(1-\epsilon)\gamma]\cdot\Ib\preceq\widehat{\Hb}(\wb)\preceq[(1+\epsilon)L]\cdot\Ib.
\end{split}
\end{equation}
}}
\end{proof}
\end{corollary}
\noindent
{\bf Discussion on efficiency of RFN vs. Newton:} Based on the result in Lemma \ref{spectral_lemma}, if $\mu=O(n^p)$ for any $0<p\leq 1$, both $\norm{\frac
{\Zb\Zb^{\top}+\mu\Ib}{n}}$ and $\norm{\frac{\Kb_1+\mu\Ib}{n}}$ are $O(1)$ since both $\norm{\Zb\Zb^{\top}}$ and $\norm{\Kb_1}$ are $O(n)$. Therefore, disregarding the log factor, the lower bound for random features (with respect to problem-dependent quantities) is $\Omega(\frac{n\norm{\Kb_1}}{\gamma^2})$. Notice that the Hessian $\Hb(\wb)$ is lower bounded as follows
\begin{equation*}
    \lambda \mu\Ib\preceq \lambda \Kb \preceq \Hb(\wb),
\end{equation*}
so letting $\gamma=\lambda\mu$, we can get $$\frac{n\norm{\Kb_1}}{\gamma^2} = \frac{n\norm{\Kb_1}}{\lambda^2\mu^2} \leq \frac{n^2}{\lambda^2\mu^2}.$$ 
Therefore, if $\mu=O(n^p)$ for any $\frac{1}{2}<p\leq 1$, the required number of random features is $o(n)$, and based on Table \ref{table:complexity}, RFN has a smaller time complexity compared to Newton method. 

% Notice that proving Lemma \ref{spectral_lemma} with standard results in the literature of kernel approximation yields weaker guarantees. In particular, using $\epsilon$-spectral approximation results (e.g., Theorem \ref{T:RFF spectral approximation}) or spectral norm errors for kernel approximation (e.g., Eq. (6.5.7) in \cite{tropp2015introduction}) give rise to $m=\Omega(\epsilon^{-2})$. However, Lemma \ref{spectral_lemma} shows that the number of required features depends on the interplay of several parameters. In particular, when the the regularization parameter $\lambda$ is small enough, $m=\Omega(\epsilon^{-1})$ since we are mainly concerned with approximating $\Kb\Db(\wb)\Kb$ in \eqref{eq:Hessian}. In particular, (disregarding the log-factors) when for a given $\epsilon>0$, we choose $\lambda^2 \leq \lambda_1\epsilon$, we have that $m=\Omega(\epsilon^{-1})$, which is a blessing obtained by regularization. On the other hand, $m=\Omega(n^2)$ suggests worse dependence on $n$ compared to Theorem \ref{T:RFF spectral approximation}, where $m=\Omega(n)$. However, this is due to the fact that we look at an error bound that depends on $\epsilon$, which is rather different from \eqref{spectral}, where the spectral inequality depends on $\epsilon \Kb$.  

% \begin{remark}
% For the rest of the paper, we assume that $\epsilon$ and $\mu$ satisfy the constraints in Lemma \ref{spectral_lemma} and Corollary \ref{Corollary: upperlower approximated Hessian}.
% \end{remark}

\subsection{Global Linear Convergence}\label{sec:global}
In this section, we establish the global convergence of our method using the approximation inequality derived in Lemma \ref{spectral_lemma}. Recall that the RFN update is written as 
\begin{equation} \label{eq:iter_update}
    \wb_{t+1} = \wb_{t} + \alpha_{t}\pb_{t},
\end{equation}
where $\pb_{t}\triangleq-[\widehat{\Hb}(\wb_{t})]^{-1}\nabla F(\wb_{t})$ and $\alpha_t$ is selected by Armijo-type line search such that
\begin{equation}\label{eq:amijo}
    F(\wb_t+\alpha_t \pb_t)\leq F(\wb_t)+\alpha_t \beta \pb_t^\top\nabla F(\wb_t),
\end{equation}
for some $\beta\in (0,1)$. In what follows, we denote
$$
\wb^*\triangleq\argmin_{\wb \in \R^n} F(\wb).
$$
\begin{theorem}\label{Global_conv}
(Global Convergence) Let Assumption \ref{A1} hold. If we update $\wb_t\in\R^n$ using RFN algorithm, where $\widehat{\Hb}(\wb_t)$ is constructed by sampling 
$$m = \Omega\left(\frac{\zeta^2n\norm{\Kb_1}}{\epsilon^2\gamma^2}\log\frac{\text{Tr}(\Kb_1)}{\norm{\Kb_1}\delta}\right),
$$
random features, we have
\begin{equation*}
    F(\wb_{t+1})-F(\wb^*)\leq (1-\rho_t)(F(\wb_t)-F(\wb^*)),
\end{equation*}
with probability at least $1-\delta$, where 
$$\rho_t\triangleq\frac{2\alpha_t\beta}{(1+\epsilon)\kappa}.$$ 
Moreover,  the step size $\alpha_t\leq 2(1-\beta)(1-\epsilon)/\kappa$ is sufficient to pass the line search. 
\end{theorem}
Theorem \ref{Global_conv} draws a connection between the precision of approximated hessian and the convergence speed. If the precision parameter $\epsilon$ is set to a small number (high precision), the step size $\alpha_t$ is upper bounded by a larger number, which implies a more aggressive update. Also, the parameter $\rho_t$ tends to be larger (i.e., faster convergence) if the approximated Hessian is closer to the original Hessian.

\subsection{Local Superlinear Convergence}\label{sec:local}
The Newton method is particularly appealing for its local convergence property, resulting in quadratic rates for strongly convex and smooth problems. 
In this section, we discuss the local convergence behavior of RFN. We use the unit step size, i.e., $\alpha_t=1$. In the following lemma, we provide an error recursion for the update $\wb_t$ using Lemma \ref{spectral_lemma}.

\begin{lemma} (Error Recursion)\label{error_recursion2}
\noindent
Let Assumptions \ref{A1}-\ref{A2} hold. If we update $\wb_t\in\R^n$ using RFN algorithm (with $\alpha_t=1)$, where $\widehat{\Hb}(\wb_t)$ is constructed by sampling 
$$m = \Omega\left(\frac{\zeta^2n\norm{\Kb_1}}{\epsilon^2\gamma^2}\log\frac{\text{Tr}(\Kb_1)}{\norm{\Kb_1}\delta}\right)
$$
random features, we have
    \begin{equation}
        \norm{\wb_{t+1}-\wb^*}\leq \nu\norm{\wb_t-\wb^*}+\eta\norm{\wb_t-\wb^*}^2,
    \end{equation}
    with probability at least $1-\delta$, where
    \begin{equation}\label{eq: Error recursion parameters}
        \nu\triangleq\frac{\epsilon}{(1-\epsilon)} ~~~~~\text{and}~~~~~  \eta\triangleq\frac{M}{2(1-\epsilon)\gamma}.  
    \end{equation}
    
\end{lemma}
The above lemma helps with establishing the local linear rate, as discussed in the following theorem.  

\begin{theorem}(Local Linear Convergence)\label{local_linear}
    Let Assumptions \ref{A1}-\ref{A2} hold. Suppose $\epsilon$ is chosen such that $0<\nu<1$ and $\rho$ is selected as 
    $$\nu<\rho<1.$$ 
    % Let $\epsilon \leq \frac{\nu\gamma+\mu\nu-\mu}{4(1+\nu)}$ and
    Assume the initial point $\wb_0$ satisfies
    \begin{equation}\label{eq:local_area}
        \norm{\wb_0-\wb^*}\leq \frac{\rho-\nu}{\eta},
    \end{equation}
    where $\eta$ is defined in Lemma \ref{error_recursion2}. If we update $\wb_t\in\R^n$ using RFN algorithm (with $\alpha_t=1)$, where $\widehat{\Hb}(\wb_t)$ is constructed by sampling 
$$m = \Omega\left(\frac{\zeta^2n\norm{\Kb_1}}{\epsilon^2\gamma^2}\log\frac{\text{Tr}(\Kb_1)}{\norm{\Kb_1}\delta}\right)
$$
random features, we have the following linear convergence
    \begin{equation}
        \norm{\wb_t-\wb^*}\leq\rho\norm{\wb_{t-1}-\wb^*},\quad t = 1,\ldots,t_0
    \end{equation}
    with probability $(1-t_0\delta)$.
\end{theorem}
While Theorem \ref{local_linear} establishes the local {\it linear} rate, we are more interested in stronger local convergence guarantees. If the precision of the Hessian approximation increases through iterations, it is expected that the algorithm can converge faster than the linear rate. In the next theorem, we show that if the Hessian approximation error decreases geometrically, RFN converges {\it superlinearly} and (asymptotically) achieves the same local rate as the Newton method.

\begin{theorem}(Local Superlinear Convergence)
    Let the assumptions of Theorem \ref{local_linear} hold. Also, for each iteration $t=0,1,\ldots,t_0$ define the quantities $\epsilon_t$, $\nu_t$ and $\eta_t$ as follows
    $$\epsilon_t=\rho^t\epsilon,~~~~~\nu_t\triangleq \frac{\epsilon_t}{(1-\epsilon_t)}~~~~~ \text{and}~~~~~
    \eta_t\triangleq\frac{M}{2(1-\epsilon_t)\gamma}.$$
    % where $\nu_t$ and $\eta_t$ are defined by setting $\mu_t=2\epsilon_t$ and substituting $\mu_t$ and $\epsilon_t$ into \eqref{eq: Error recursion parameters}. 
    Assume that $\wb_0$ satisfies \eqref{eq:local_area} with $\rho$, $\nu_0$ and $\eta_0$. Then, RFN algorithm (with $\alpha_t=1)$ achieves the following superlinear convergence
    \begin{equation}\label{eq: super linear convergence}
        \norm{\wb_t-\wb^*}\leq \rho^t\norm{\wb_{t-1}-\wb^*},\quad t=1,\ldots,t_0,
    \end{equation}
    with probability at least $(1-t_0\delta)$.
\end{theorem}
\begin{proof}
The proof follows a similar argument to that of Theorem 7 in \cite{roosta2019sub}. Based on Lemma \ref{error_recursion2}, for each iteration $t$, sampling $m$ random features such that the Hessian approximation error is at most $\epsilon_t$, we have
$$\norm{\wb_{t+1}-\wb^*}\leq \nu_t\norm{\wb_t-\wb^*}+\eta_t\norm{\wb_t-\wb^*}^2.$$
Note that, by $\epsilon_t=\rho^t\epsilon$, it follows that
$$\nu_t\leq \rho^t\nu_0$$
$$\eta_t\leq \eta_{t-1}.$$ 
Define $\Delta_t \triangleq \wb_t-\wb^*$. For $t=0$, by assumption on $\rho$, $\nu_0$, and $\eta_0$ (Theorem \ref{local_linear}), we have 
\begin{eqnarray*}
    \norm{\Delta_1} &\leq& \nu_0\norm{\Delta_0} + \eta_0\norm{\Delta_0}^2\\
    &\leq& \rho\norm{\Delta_0}.
\end{eqnarray*}
Assume \eqref{eq: super linear convergence} holds up to iteration $t$. For $t+1$, we get
\begin{eqnarray*}
    \norm{\Delta_{t+1}} &\leq& \nu_t\norm{\Delta_t} + \eta_t\norm{\Delta_t}^2\\
    &\leq& \rho^t\nu_0\norm{\Delta_t} + \eta_t\norm{\Delta_t}^2\\
    &\leq&\rho^t\nu_0\norm{\Delta_t} + \eta_0\norm{\Delta_t}^2.
\end{eqnarray*}
By induction hypothesis, we have $\norm{\Delta_{t-1}}\leq \norm{\Delta_0}$, and 
$$\norm{\Delta_{t}}\leq \rho^t\norm{\Delta_{t-1}}\leq \rho^t\big(\frac{\rho-\nu_0}{\eta_0}\big).$$
Therefore, it follows that $\norm{\Delta_{t+1}}\leq \rho^{t+1}\norm{\Delta_{t}}$.
\end{proof}

\begin{figure*}[ht!] 
    \centering
    \includegraphics[width=0.65\columnwidth]{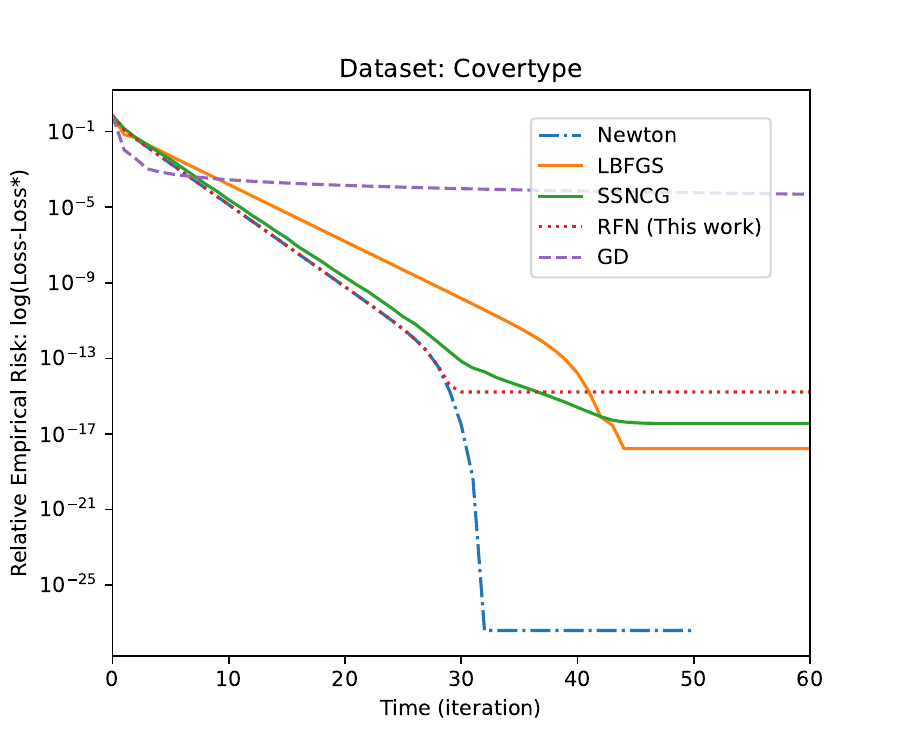}
    \includegraphics[width=0.65\columnwidth]{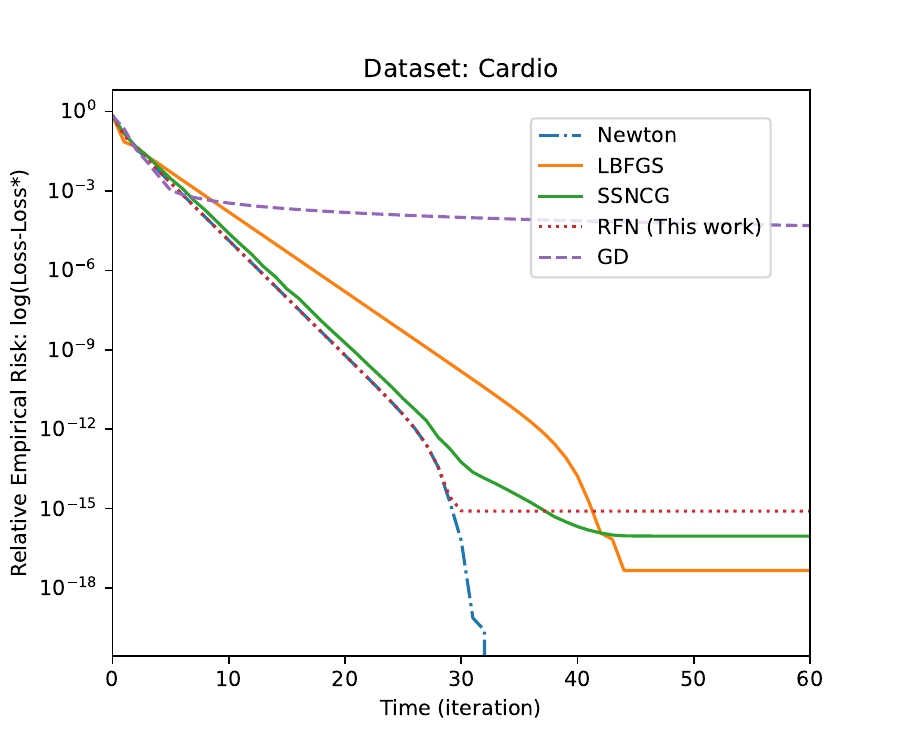}
    \includegraphics[width=0.65\columnwidth]{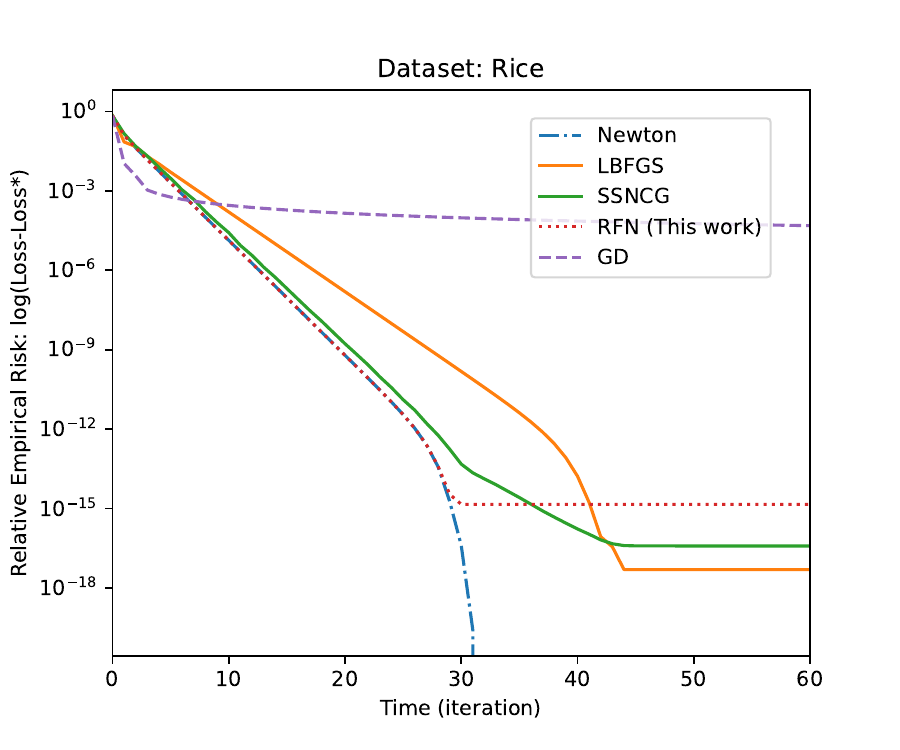}
    \caption{The plot of relative empirical risk (training error) vs. iteration shows that RFN enjoys a similar iteration complexity compared to the original Newton method. In this sense, RFN outperforms L-BFGS, SSNCG, and GD.}
    \label{fig:Results}
\end{figure*}

\section{Numerical Experiments}\label{sec:simul}
We now provide numerical experiments to illustrate the performance of the random-feature based Newton method. We consider regularized kernel logistic regression 
\begin{equation}\label{eq: loss function experiment}
    \resizebox{1\hsize}{!}{%
    $F(\wb)=\frac{1}{n}\sum\limits_{i=1}^n\log(1+\exp(-y_i\sum\limits_{j=1}^nk(\xb_i,\xb_j)w_j))+\lambda \norm{\wb}_{\Kb}^2$,
    }
\end{equation}
where $\Kb= \Kb_1 + \mu \Ib$ is a composite positive-definite kernel, and $\norm{\wb}_{\Kb}^2=\wb^\top\Kb\wb$ is the norm with respect to the gram matrix. We choose $k_1(\xb,\xb')=\exp(-\norm{\xb-\xb'}^2/2\sigma^2)$ to be a Gaussian kernel. The goal is to apply the random feature technique to approximate the Gaussian kernel.

\noindent
{\bf -- Benchmark algorithms:} 
We compare RFN (as described in Algorithm \ref{alg:RFN}) with the algorithms listed below:
\begin{enumerate}
    %\item {\bf RFN:} {\rd The random-feature based Newton method $\wb_{t+1} = \wb_{t}+\alpha_t\pb_t$ where $\pb_t=-\widehat{\Hb}^{-1}(\wb_t)\nabla F(\wb_t)$. $\widehat{\Hb}$ is derived by approximating $\Kb_1$ with random features, and $\widehat{\Hb}^{-1}$ is computed by applying the matrix lemma. For the sake of efficiency, the set of random features is sampled in the first iteration and re-used for the rest update process.}
    \item {\bf Newton:} The original Newton method $\wb_{t+1} = \wb_{t}+\alpha_t\pb_t$, where $\pb_t$ is the exact solution of the linear system $\Hb(\wb_t)\pb_t=-\nabla F(\wb_t)$.
    \item {\bf SSNCG:} The sub-sampled Newton method with conjugate gradient update $\wb_{t+1} = \wb_{t}+\alpha_t\pb_t$, where $\pb_t$ is computed by solving the linear system $\widehat{\Hb}_{\text{SSN}}(\wb_t)\pb_t=-\nabla F(\wb_t)$ up to a high precision using the CG method. The approximated Hessian $\widehat{\Hb}_{\text{SSN}}(\wb_t)$ is generated based on the description in Section \ref{sec:ssn}, and the Hessian is approximated as follows 
    \begin{align*}
    \widehat{\Hb}_{\text{SSN}}(\wb)&=\sum_{i\in \Ic}[\Db(\wb)]_{ii}\Kb(\Vc,i)\Kb(i,\Vc)\\
    &+\lambda\big(\Kb_1(\Vc,\Ic)\Kb_1(\Ic,\Ic)^{\dagger}\Kb_1(\Ic,\Vc) + \mu\Ib\big),
    \end{align*}
    where $\Ic$ is a random subset of $\{1,\ldots,n\}$. SSNCG is the adaption of algorithms in \cite{bollapragada2018exact,roosta2019sub} as described in Section \ref{sec:ssn}.
    \item {\bf L-BFGS:} Limited-memory BFGS which approximates the Newton direction using the first order information. Here, we implement BFGS using the history of the past 50 updates of $\wb_t$ and $\nabla F(\wb_t)$.
    \item {\bf GD:} The plain gradient descent method where $\pb_t = -\nabla\Fb(\wb_t)$.
\end{enumerate}
For all these methods, the step size $\alpha_t$ is determined by Armijo backtracking line search, and the full gradient $\nabla F(\wb_t)$ is used. For SSNCG, the linear system is solved approximately to achieve $10^{-6}$ relative error.

\noindent
{\bf -- Hyper-parameters of the empirical risk:} There are three hyper-parameters in the regularized kernel logistic regression with a composite kernel: $\sigma$, $\mu$ and $\lambda$. $\lambda$ determines the rate by which we impose the RKHS norm as a penalty function. When $\lambda$ is small, the model fits the training data. On the other hand, $\sigma$, known as the bandwidth of a Gaussian kernel, determines the kernel generalization characteristics. When $\sigma$ is small, the gram matrix gets closer to a full-rank matrix, and the corresponding distribution of random features has a larger variance, which requires more random feature samples to approximate the kernel function. However, for a Gaussian kernel with a large bandwidth, the gram matrix eigenvalues decay faster, and $\epsilon$ has to be very small to ensure the positive definiteness of the approximated Hessian (see  \eqref{eq:upperlower}). $\mu$ controls the relative importance of $k_1(\xb,\xb')$ and as $\mu$ grows larger, the Gaussian kernel plays a less important role for the composite kernel.

% \noindent
% {\bf -- Hyper-parameter of backtracking line search:} We use a simple grid search to optimize the parameters for Newton and L-BFGS. However, due to the randomness of the approximated Hessian for randomized methods, we perform the line search adaptively for both SSNCG and RFN. For each iteration, the initial step size of backtracking is set as twice of the step size chosen in the previous iteration. This allows the algorithms to avoid staying in the search loop for long. \\

\begin{figure*}[ht!] 
    \centering
    \includegraphics[width=0.65\columnwidth]{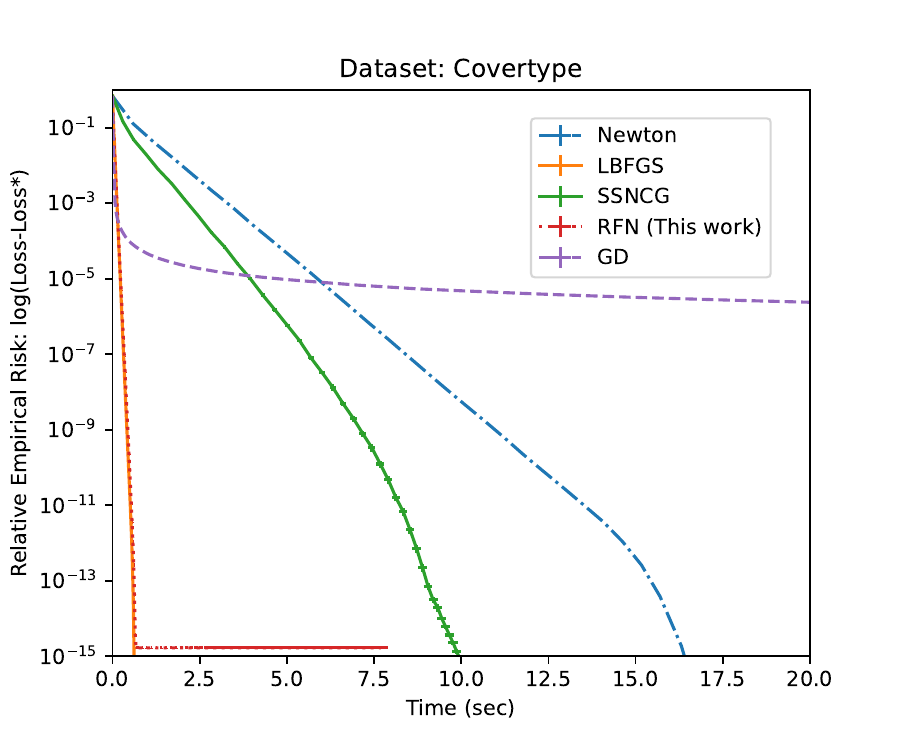}
    \includegraphics[width=0.65\columnwidth]{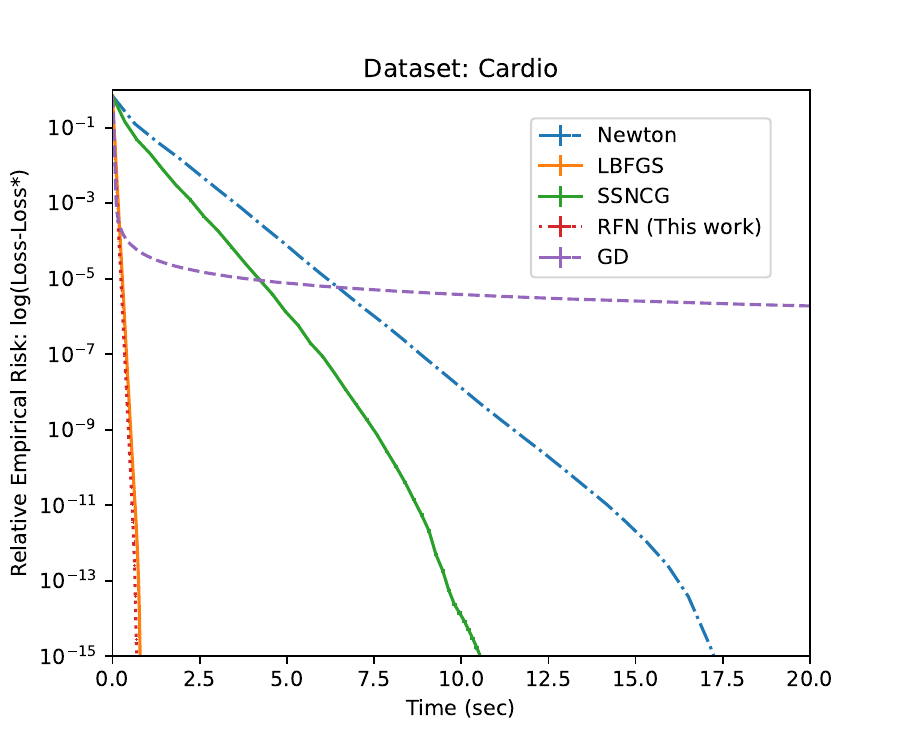}
    \includegraphics[width=0.65\columnwidth]{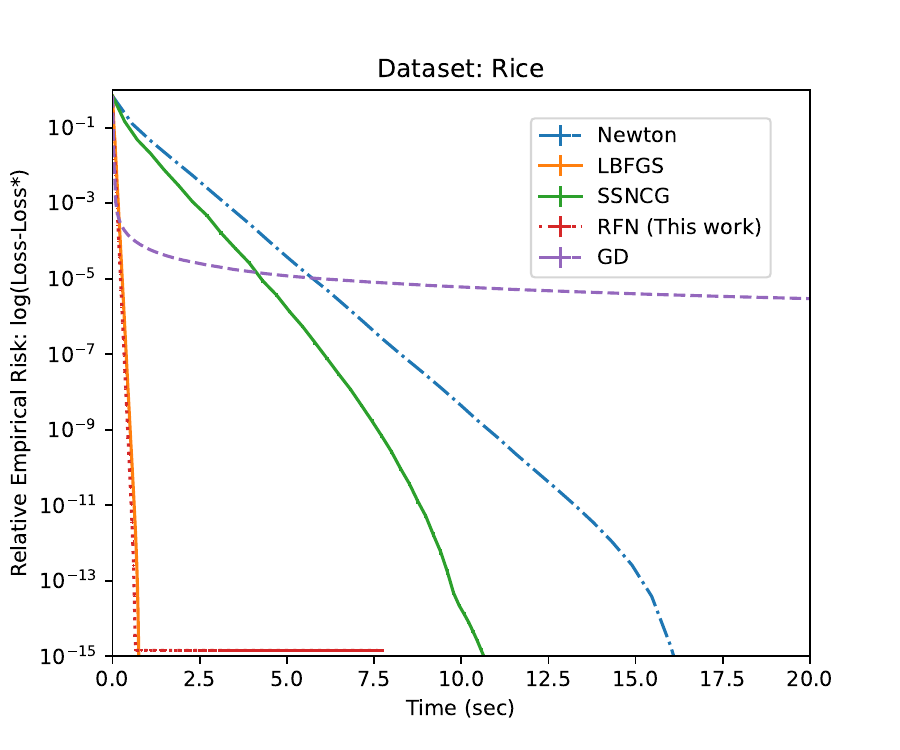}
    \caption{The plot of relative empirical risk (training error) vs. time cost shows that RFN outperforms nearly all other methods. Only L-BFGS has a similar performance to RFN.}
    \label{fig:Results2}
\end{figure*}

\noindent
{\bf -- Datasets:} We apply all of the methods on three datasets from the UCI Machine Learning Repository and Kaggle (Table \ref{table:dataset}). For each dataset, we randomly sample $n=3000$ data points for training to run the optimization. 
\begin{table}[t!]
\caption{Description of datasets used in the experiments.}
\centering
\begin{adjustbox}{width=.95\columnwidth}
\begin{tabular}{||c c c c||} 
 \hline
 Dataset & \# of points & \# of features & Reference \\ [0.5ex] 
 \hline\hline
 Covertype & 581012 & 54 & \cite{blackard1999comparative} \\ 
%  Cod-RNA & 59535 & 8 & \cite{uzilov2006detection} \\
 Cardio & 70000 & 11 & \cite{UlianovaCardio}\\
 Rice & 18185 & 10 & \cite{MsSmartyPantsRice}\\
 \hline
\end{tabular}
\end{adjustbox}
\label{table:dataset}
\end{table}

Based on the data distribution of each dataset, the hyper-parameters  are set as follows:\\
{\it Covertype:} $\sigma^2=5$, $\lambda=2*10^{-15}$, $\mu=1000$, $\alpha=0.3$ and $\beta=0.5$.\\
{\it Cardio:} $\sigma^2=100$, $\lambda=2*10^{-15}$, $\mu=1000$, $\alpha=0.3$ and $\beta=0.5$.\\
{\it Rice:} $\sigma^2=50$, $\lambda=2*10^{-15}$, $\mu=1000$, $\alpha=0.3$ and $\beta=0.5.$

For the number of random features for RFN ($m$) and the number of sub-sampled data points for SSNCG ($|\Ic|$), we set the ratio as $10\%$ of the training data (i.e., $m=|\Ic|=300$). 

\noindent
{\bf -- Performance:} We record the loss value and the time cost of each iteration for all methods. The time cost includes the time of finding the Newton direction (computing the Hessian and solving the linear system) and determining the step size. The run time is based on a desktop with a 6-core, AMD Ryzen 5 5600G CPU  and 15.5G of RAM (3600Mhz). The initial point $\wb_0$ is the all zero vector. For randomized methods RFN and SSNCG, we run the experiment 30 times and report the average both for the loss value and time cost.   

Based on Fig. \ref{fig:Results}, we see that in terms of iteration complexity, RFN and SSNCG outperform GD and L-BFGS, which leverage first order information to approximate the Hessian. From Fig. \ref{fig:Results2}, we observe that RFN and L-BFGS have the most competitive loss vs. run time performance. The two figures together imply that RFN achieves an identical iteration complexity to the original Newton method with a cheap computational cost. In this simulation, for RFN we compute $\Db(\wb)$ and the gradient based on the approximated gram matrix, which leads to the saturation observed in Fig. \ref{fig:Results}.

\begin{figure}[t!]
    \centering
    \includegraphics[width=0.8\columnwidth]{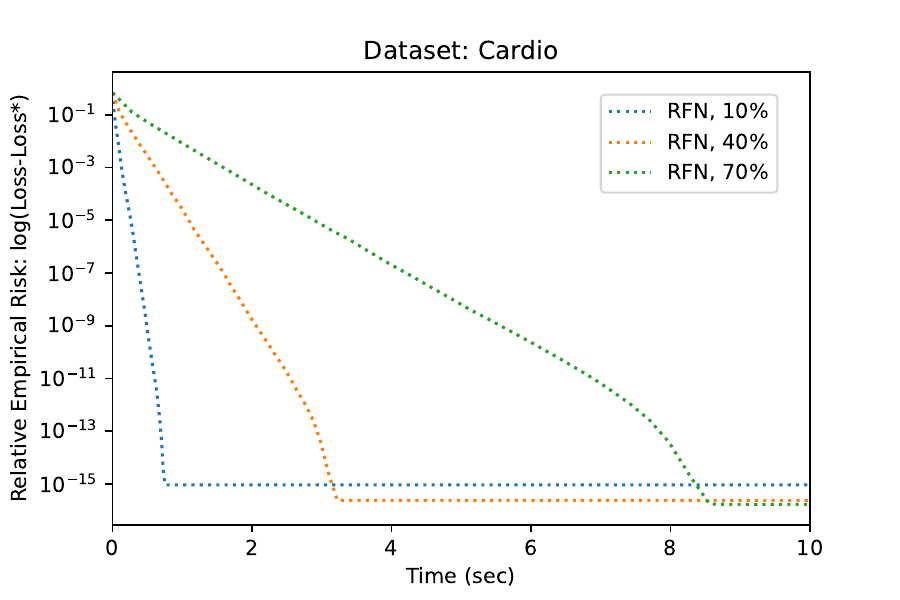}
    \caption{Empirical risk vs. time cost for different ratios of $m/n$.}
    \label{fig:ratio_time}
\end{figure}
\begin{figure}[t!] 
    \centering
    \includegraphics[width=0.8\columnwidth]{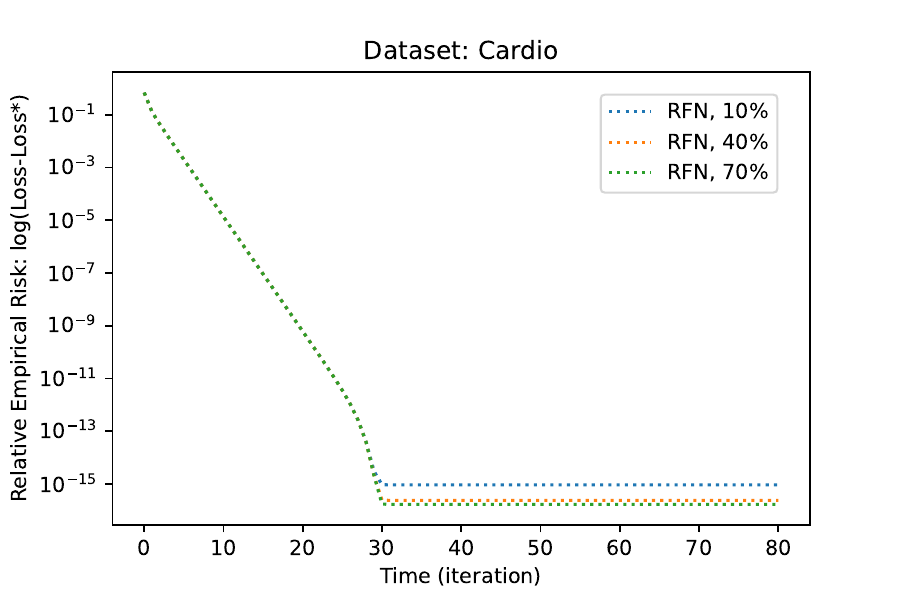}
    \caption{Empirical risk vs. iterations for different ratios of $m/n$.}
    \label{fig:ratio_iteration}
\end{figure}

\noindent
{\bf -- Impact of $m$:} Another factor affecting the convergence performance of RFN is the number of random features. 
We run another experiment on Cardio dataset with the same hyper-parameter set up except that we choose the ratio ($m/n$) from the set $\{10\%, 40\%, 70\%\}$. Figs. \ref{fig:ratio_time}-\ref{fig:ratio_iteration} verify that with more random features, we can achieve a smaller loss, but this comes at the cost of increased run time. The reason is that using more random features improves the Hessian approximation, but it also increases 
the time of computing the newton step based on \eqref{eq:hessianIL2}.

\section{Conclusion}
In this paper, we proposed a random-feature based Newton method (RFN) for risk minimization over RKHS. We drew explicit connections between the Hessian and the gram matrix and observed that sub-sampled Newton methods are not directly applicable to this optimization problem. Then, we showed that the Newton method can be expedited by applying kernel approximation techniques. From the theoretical point of view, we proved that the approximate Hessian is close to the original Hessian in terms of spectral norm when enough random features are sampled, which in turn ensures the local and global convergence with high probability. From the practical point of view, we applied RFN to three real-world datasets, compared it with other benchmarks, and showed that RFN enjoys a faster run time under certain conditions. Future directions include the development of distributed/decentralized variants of RFN as well as data-dependent sampling schemes for random features \cite{wang2021orcca}.

\section{Appendix}\label{sec:appendix}
We make use of the following matrix concentration inequality \cite{tropp2015introduction} for the proof of Lemma \ref{spectral_lemma}, but we adopt the version used in \cite{avron2017random} and present it for real matrices.

\noindent
\begin{lemma}\label{matrix_concentration}(Matrix Concentration Inequality) Let $\Bb$ be a fixed $d_1 \times d_2$ matrix. Consider a $d_1 \times d_2$ random matrix $\Rb$ that satisfies 
$$\mathbb{E}[\Rb] = \Bb\quad\text{and}\quad\norm{\Rb}\leq U.$$
Let $\Mb_1$ and $\Mb_2$ be semi-definite upper bounds such that
$$\mathbb{E}[\Rb\Rb^\top]\preceq \Mb_1\quad\text{and}\quad\mathbb{E}[\Rb^\top\Rb]\preceq \Mb_2.$$
Define the quantities
$$c\triangleq\max(\norm{\Mb_1}, \norm{\Mb_2})\quad\text{and}\quad  b\triangleq(\text{Tr}(\Mb_1)+\text{Tr}(\Mb_2))/c.$$ 
Form the matrix sampling estimator
$$\Bar{\Rb}_h = \frac{1}{h}\sum_{i=1}^h\Rb_i,$$
where each $\Rb_i$ is an independent copy of $\Rb$. Then, for all $\epsilon\geq \sqrt{c/h}+2U/3h,$
$$\text{Pr}\left(\norm{\Bar{\Rb}_h-\Bb}\geq \epsilon\right)\leq 4b\exp{\Big(\frac{-h\epsilon^2/2}{c+2U\epsilon/3}\Big)}.$$
\end{lemma}

\subsection{Proof of Lemma \ref{spectral_lemma}}
In the proof of this lemma, we disregard the dependence on $\wb$ and denote $\widehat{\Hb}(\wb)$, $\Hb(\wb)$ and $\Db(\wb)$ by $\widehat{\Hb}$, $\Hb$ and $\Db$, respectively. The introduced shorthand is just for the sake of presentation clarity.

First, we focus on $\norm{\Zb\Zb^\top-\Kb_1}$, which can be written as $\norm{\frac{1}{m}\sum_{i=1}^m \zb_i\zb_i^\top-\Kb_1}$, where $\zb_i$ is the $i$-th column of $\Zb$. Notice that $\mathbb{E}[\zb_i\zb_i^\top]=\Kb_1$. To apply Lemma \ref{matrix_concentration}, we observe that $$\norm{\zb_i\zb_i^\top}\leq n\triangleq U,$$
and
\begin{align*}
    \E[\zb_i\zb_i^\top\zb_i\zb_i^\top]\preceq n\Kb_1\triangleq \Mb_1.
\end{align*}
By symmetry $\Mb_1=\Mb_2$, and we also have that
$$c=n\norm{\Kb_1}$$
$$b=\frac{2\text{Tr}(\Mb_1)}{c}=\frac{2\text{Tr}(\Kb_1)}{\norm{\Kb_1}}.$$
With the above quantities, based on Lemma \ref{matrix_concentration}, we derive
\begin{align*}
    &\text{Pr}\left(\norm{\frac{1}{m}\sum_{i=1}^m \zb_i\zb_i^\top-\Kb_1}\geq\epsilon \right)
    \leq 4b\exp\Big(\frac{-m\epsilon^2 /2}{n\norm{\Kb_1}+\frac{2}{3}n\epsilon }\Big).
\end{align*}
To have the right hand side smaller than $\delta$, we should solve the inequality
$$\log\frac{4b}{\delta}\leq \frac{m\epsilon^2/2}{n\norm{\Kb_1}+\frac{2}{3}n\epsilon},$$
for $m$, which implies
$$
    m\geq \frac{2n\norm{\Kb_1}+\frac{4}{3}n\epsilon }{\epsilon^2}\log\frac{4b}{\delta}.
$$
Since we further assume that $\epsilon\leq \frac{3}{2} \norm{\Kb_1}$, the sufficient lower bound on $m$ can be simplified to
\begin{equation}\label{eq: gram matrix sample complexity lower bound}
    m\geq\frac{4n\norm{\Kb_1}}{\epsilon^2}\log\frac{4b}{\delta}.
\end{equation}
Therefore, by sampling $m = \Omega(\frac{n\norm{\Kb_1}}{\epsilon^2}\log\frac{\text{Tr}(\Kb_1)}{\norm{\Kb_1}\delta})$, with probability at least $(1-\delta)$, we have
\begin{equation}\label{eq: gram matrix diff bound}
    \norm{\Zb\Zb^{\top} - \Kb_1} \leq \epsilon.
\end{equation}
Denoting $(\Zb\Zb^{\top} + \mu\Ib)$ as $\widehat{\Kb}$, based on \eqref{eq: gram matrix diff bound} the approximation error of Hessian is upper-bounded as follows:
\begin{equation}\label{eq2: hessian Approximation}
\begin{split}
    \norm{\widehat{\Hb} - \Hb}=&\norm{\frac{\widehat{\Kb}\Db\widehat{\Kb}}{n} + \lambda \widehat{\Kb} - \frac{\Kb\Db\Kb}{n} - \lambda \Kb}\\
    \leq &\norm{\frac{\widehat{\Kb}\Db\widehat{\Kb} - \Kb\Db\Kb}{n} } + \norm{\lambda (\Zb\Zb^{\top} - \Kb_1)}\\
    \leq &\norm{\frac{\widehat{\Kb}\Db\widehat{\Kb} - \widehat{\Kb}\Db\Kb}{n} }\\
    + &\norm{\frac{\widehat{\Kb}\Db\Kb - \Kb\Db\Kb}{n} } + \norm{\lambda (\Zb\Zb^{\top} - \Kb_1)}\\
    \leq &\left(\lambda + \norm{\frac{\widehat{\Kb}\Db}{n}} + \norm{\frac{\Db\Kb}{n}}\right)\norm{\Zb\Zb^{\top} - \Kb_1}\\
    \leq &\left(\lambda + \lambda_1\norm{\frac{\widehat{\Kb}}{n}} + \lambda_1\norm{\frac{\Kb}{n}}\right)\epsilon.
\end{split}
\end{equation}
Define $\zeta \triangleq \lambda + \lambda_1\norm{\frac{\widehat{\Kb}}{n}} + \lambda_1\norm{\frac{\Kb}{n}}$. Based on \eqref{eq2: hessian Approximation}, by replacing $\epsilon$ with $(\epsilon\gamma)/\zeta$ in \eqref{eq: gram matrix sample complexity lower bound}, the result is proved.

\subsection{Proof of Theorem \ref{Global_conv}}
Denote the minimum eigenvalue of $\widehat{\Hb}(\wb_t)$ by $\lambda_{\text{min}}(\widehat{\Hb}(\wb_t))$, and recall that $\pb_{t}\triangleq-[\widehat{\Hb}(\wb_{t})]^{-1}\nabla F(\wb_{t})$. Since $$\pb_t^\top\widehat{\Hb}(\wb_t)\pb_t\geq \lambda_{\text{min}}(\widehat{\Hb}(\wb_t))\norm{\pb_t}^2,$$
by Corollary \ref{Corollary: upperlower approximated hessian} we have that 
$$\pb_t^\top\widehat{\Hb}(\wb_t)\pb_t\geq (1-\epsilon)\gamma \norm{\pb_t}^2.$$
For any $\alpha>0$, define $\wb_{\alpha}\triangleq\wb_t+\alpha \pb_t$. With $L$-smoothness of the objective function $F$, we have
\begin{eqnarray*}
    &&F(\wb_{\alpha})-F(\wb_t)\\&\leq&(\wb_{\alpha}-\wb_t)^\top\nabla F(\wb_t)+\frac{L}{2}\norm{\wb_{\alpha}-\wb_t}^2\\&=&\alpha \pb_t^\top\nabla F(\wb_t)+\frac{\alpha^2L}{2}\norm{\pb_t}^2.
\end{eqnarray*}
To pass the Armijo line search condition, we need an $\alpha$ which makes the following inequality hold
$$\alpha \pb_t^\top\nabla F(\wb_t)+\frac{\alpha^2L}{2}\norm{\pb_t}^2\leq\alpha \beta \pb_t^\top\nabla F(\wb_t).$$
Since $\pb_t^\top\nabla F(\wb_t)=-\pb_t^\top\widehat{\Hb}(\wb_t)\pb_t$, the above inequality can be written as 
$$\alpha L\norm{\pb_t}^2\leq 2(1-\beta)\pb_t^\top\widehat{\Hb}(\wb_t)\pb_t.$$
Therefore, if 
$$\alpha\leq 2(1-\beta)(1-\epsilon)\gamma/L,$$
(given that $\pb_t^\top\widehat{\Hb}(\wb_t)\pb_t\geq (1-\epsilon)\gamma\norm{\pb_t}^2$), the Armijo line search is satisfied. This upper bound is iteration-independent. Now, $\wb_{t+1}=\wb_t + \alpha_t\pb_t$, and based on (\ref{eq:amijo}), we have that
\begin{eqnarray*}
  \vphantom{\alpha_t\beta\frac{\norm{\nabla F(\wb_t)}^2}{[(1+\Psi)L+\mu]}}F(\wb_{t+1})&\leq& F(\wb_t) + \alpha_t\beta\pb_t^\top\nabla F(\wb_t)\\
  \vphantom{\alpha_t\beta\frac{\norm{\nabla F(\wb_t)}^2}{[(1+\Psi)L+\mu]}}&=&F(\wb_t) - \alpha_t\beta\nabla F(\wb_t)^\top[\widehat{\Hb}(\wb_t)]^{-1}\nabla F(\wb_t)\\
  &\leq&F(\wb_t) - \alpha_t\beta\frac{\norm{\nabla F(\wb_t)}^2}{(1+\epsilon)L},
\end{eqnarray*}
where the last inequality comes from \eqref{eq:upperlower}. 
By subtracting $F(\wb^*)$ from both sides and noting that $F(\wb_t)-F(\wb^*)\leq\frac{\norm{\nabla F(\wb_t)}^2}{2\gamma}$ due to the strong convexity of $F$, the result is proved.

\begin{lemma}\label{error_recursion} (Error Recursion)
    Let Assumption \ref{A2} hold. Assume that $\alpha_t=1$ and $\widehat{\Hb}(\wb)$ is positive definite. We then have
    $$\norm{\wb_{t+1}-\wb^*}\leq \nu\norm{\wb_t-\wb^*}+\eta \norm{\wb_t-\wb^*}^2,$$
    where $\eta\triangleq\frac{M}{2\lambda_{\text{min}}(\widehat{\Hb}(\wb_t))}$ and $\nu\triangleq\frac{\norm{\widehat{\Hb}(\wb_t)-\Hb(\wb_t)}}{\lambda_{\text{min}}(\widehat{\Hb}(\wb_t))}$.
\end{lemma}

\subsection{Proof of Lemma \ref{error_recursion}}
Define $\Delta_t\triangleq\wb_t-\wb^*$. Since $$\wb_{t+1}=\wb_t-\widehat{\Hb}(\wb_t)^{-1}\nabla F(\wb_t),$$ 
$\wb_{t+1}$ is the optimal solution of the following second order approximation: 
\begin{align*}
    N(\wb)&\triangleq F(\wb_t)+(\wb-\wb_t)^\top\nabla F(\wb_t)\\
    &+\frac{1}{2}(\wb-\wb_t)^\top\widehat{\Hb}(\wb_t)(\wb-\wb_t).
\end{align*}
Based on that, for any $\wb\in\R^n$, we have
\begin{align*}
0=&(\wb-\wb_{t+1})^\top\nabla N(\wb_{t+1})\\=&(\wb-\wb_{t+1})^\top\nabla F(\wb_t)+(\wb-\wb_{t+1})^\top\widehat{\Hb}(\wb_t)(\wb_{t+1}-\wb_t).
\end{align*}
By setting $\wb=\wb^*$ and noting that $\wb_{t+1}-\wb_t=\Delta_{t+1}-\Delta_t$, we have 
\begin{align*}
\Delta_{t+1}^\top\widehat{\Hb}(\wb_t)\Delta_{t+1}&= \Delta_{t+1}^\top\widehat{\Hb}(\wb_t)\Delta_{t}-\Delta_{t+1}^\top\nabla F(\wb_t)\\
&+\Delta_{t+1}^\top \nabla F(\wb^*),
\end{align*}
as $\nabla F(\wb^*)=0$ due to the optimality of $\wb^*$. 
%$$^\top (\wb_{t+1}-\wb^*)\geq 0$$, 
%which is added to the right hand side of the above inequality and get $\Delta_{t+1}^\top\widehat{\Hb}(\wb_t)\Delta_{t+1}\leq \Delta_{t+1}^\top\widehat{\Hb}(\wb_t)\Delta_{t}-\Delta_{t+1}^\top\nabla F(\wb_t)+$. 
We also have that
\begin{eqnarray*}
    &&\nabla F(\wb_t)-\nabla F(\wb^*)\\
    &=&\left(\int_0^1 \nabla^2 F(\wb^*+\tau (\wb_t-\wb^*))d\tau\right)(\wb_t-\wb^*).
\end{eqnarray*}
Therefore, 

\scalebox{.88}{\parbox{\columnwidth}{
\begin{align*}
    \vphantom{\sum\limits_{n}}&\Delta_{t+1}^\top\widehat{\Hb}(\wb_t)\Delta_{t+1}\\
    \vphantom{\sum\limits_{n}}=&\Delta_{t+1}^\top\widehat{\Hb}(\wb_t)\Delta_{t}-\Delta_{t+1}^\top\left(\int_0^1 \nabla^2 F(\wb^*+\tau (\wb_t-\wb^*))d\tau\right)\Delta_t\\
    \vphantom{\sum\limits_{n}}=&\Delta_{t+1}^\top\widehat{\Hb}(\wb_t)\Delta_{t}-\Delta_{t+1}^\top\nabla^2 F(\wb_t)\Delta_t\\
    \vphantom{\sum\limits_{n}}+&\Delta_{t+1}^\top\nabla^2 F(\wb_t)\Delta_t-\Delta_{t+1}^\top\left(\int_0^1 \nabla^2 F(\wb^*+\tau (\wb_t-\wb^*))d\tau\right)\Delta_t\\
    \vphantom{\sum\limits_{n}}\leq&\norm{\Delta_{t+1}}\norm{\widehat{\Hb}(\wb_t)-\nabla^2 F(\wb_t)}\norm{\Delta_t}\\
    \vphantom{\sum\limits_{n}}+&\norm{\Delta_{t+1}}\left(\int_0^1\norm{\nabla^2 F(\wb_t)-\nabla^2 F(\wb^*+\tau(\wb_t-\wb^*))}d\tau\right)\norm{\Delta_{t}}\\
    \vphantom{\sum\limits_{n}}\leq&\norm{\widehat{\Hb}(\wb_t)-\nabla^2 F(\wb_t)}\norm{\Delta_t}\norm{\Delta_{t+1}}+\frac{M}{2}\norm{\Delta_t}^2\norm{\Delta_{t+1}}.
\end{align*}
}}
Since $\Delta_{t+1}^\top\widehat{\Hb}(\wb_t)\Delta_{t+1}\geq \lambda_{\text{min}}(\widehat{\Hb}(\wb_t))\norm{\Delta_{t+1}}^2$ and $\widehat{\Hb}(\wb_t)$ is assumed to be positive definite, the result follows.

\subsection{Proof of Lemma \ref{error_recursion2}}
In view of Lemma \ref{spectral_lemma} and Corollary \ref{Corollary: upperlower approximated hessian}, we know that $\norm{\widehat{\Hb}(\wb_t)-\nabla^2 F(\wb_t)}\leq \epsilon\gamma$ and $\lambda_{\text{min}}(\widehat{\Hb}(\wb_t))\geq (1-\epsilon)\gamma$. The result follows by applying these to Lemma \ref{error_recursion}.

\subsection{Proof of Theorem \ref{local_linear}}
%By choosing $\epsilon$ such that $0<\nu<1$, 
According to Lemma \ref{error_recursion2}, we have $\norm{\wb_{t+1}-\wb^*}\leq \nu\norm{\wb_t-\wb^*}+\eta\norm{\wb_t-\wb^*}^2$ for every $t$. The choice of $\wb_0$ guarantees that
$$\nu\norm{\wb_0-\wb^*}+\eta\norm{\wb_0-\wb^*}^2\leq\rho\norm{\wb_0-\wb^*},$$ 
and the proof follows by induction. 

The probability that all iterations are successful is the complement of the probability that at least one iteration fails, which is bounded by $t_0\delta$. Therefore, the probability of a successful process is at least $(1-t_0\delta)$.

\bibliographystyle{IEEEtranN}
\bibliography{references}

\end{document}